\documentclass[review]{elsarticle}
\usepackage{amssymb}
\usepackage{amsmath}
\usepackage{booktabs}
\usepackage{multirow,multicol}
\usepackage{booktabs}  
\usepackage{threeparttable} 
\usepackage{multirow}
\usepackage{color}
\usepackage{hyperref}
\usepackage{graphicx}
\usepackage[table]{xcolor}
\usepackage{ulem}
\usepackage{caption}

\usepackage{stfloats}
\usepackage{makecell}
\usepackage{colortbl}
\usepackage{algorithmic}
\usepackage{algorithm}
\usepackage{pifont}
\usepackage{subfigure}
\usepackage{cleveref}
\usepackage{placeins}
\usepackage{caption}

\journal{Nuclear Physics B}

\begin{document}

\begin{frontmatter}



\title{Action Hints: Semantic Typicality and Context Uniqueness for Generalizable Skeleton-based Video Anomaly Detection}


% \author{Canhui Tang, 
%        Sanping Zhou,
%        Haoyue Shi, Le Wang}%% Author name

% \affiliation{organization={},%Department and Organization

\author[1,2,3]{Canhui Tang}
\ead{2193512014@stu.xjtu.edu.cn}
\author[1,2,3]{Sanping~Zhou\corref{mycorrespondingauthor}}
\cortext[mycorrespondingauthor]{Corresponding author: spzhou@xjtu.edu.cn}
\author[4]{Haoyue Shi}
\ead{haoyueshi@chd.edu.cn}
\author[1,2,3]{Le~Wang}
\ead{lewang@xjtu.edu.cn}

\address[1]{National Key Laboratory of Human-Machine Hybrid Augmented Intelligence}
\address[2]{National Engineering Research Center for Visual Information and Applications}
\address[3]{Institute of Artificial Intelligence and Robotics, Xi'an Jiaotong University}
\address[4]{School of Electronics and Control Engineering, Chang'an University}

%% Abstract
\begin{abstract}
Zero-Shot Video Anomaly Detection (ZS-VAD) requires temporally localizing anomalies without target domain training data, which is a crucial task due to various practical concerns, \textit{e.g.}, data privacy or new surveillance deployments. Skeleton-based approach has inherent generalizable advantages in achieving ZS-VAD as it eliminates domain disparities both in background and human appearance. However, existing methods only learn low-level skeleton representation and rely on the domain-limited normality boundary, which cannot generalize well to new scenes with different normal and abnormal behavior patterns. In this paper, we propose a novel zero-shot video anomaly detection framework, unlocking the potential of skeleton data via action typicality and uniqueness learning. Firstly, we introduce a language-guided semantic typicality modeling module that projects skeleton snippets into action semantic space and distills LLM's knowledge of typical normal and abnormal behaviors during training. Secondly, we propose a test-time context uniqueness analysis module to finely analyze the spatio-temporal differences between skeleton snippets and then derive scene-adaptive boundaries. Without using any training samples from the target domain, our method achieves state-of-the-art results against skeleton-based methods on four large-scale VAD datasets: ShanghaiTech, UBnormal, NWPU, and UCF-Crime, featuring over 100 unseen surveillance scenes.
\end{abstract}


% \begin{highlights}
% \item We propose a skeleton-based video anomaly detection framework that learns action typicality and uniqueness, enhancing generalizability across diverse unseen target scenes.
% \item We propose a language-guided typicality modeling module that projects skeleton snippets into a generalizable semantic space and distills LLM's knowledge of typical normal and abnormal behaviors during training. 
% \item We propose a test-time uniqueness analysis module to finely analyze the spatio-temporal differences between skeleton snippets and derive scene-adaptive boundaries between normal and abnormal behavior.
% \end{highlights}

%% Keywords

\end{frontmatter}

\section{Introduction}
\label{sec:intro}

Video Anomaly Detection (VAD) aims to temporally locate abnormal events, which has wide applications in the context of video surveillance and public safety~\cite{shanghaitech,ucf}. Current mainstream paradigms include one-class~\cite{jigsaw,hf2,tang2024advancing,stgnf,wang2023memory} and weakly supervised methods~\cite{ucf,lookaround,shi2023abnormal}, which require abundant samples from the target video domain for training. However, in surveillance scenarios involving \textit{privacy} or \textit{newly installed monitoring devices}, training samples from the target domain are usually not available. Therefore, designing a Zero-Shot Video Anomaly Detection (ZS-VAD) method that can generalize to diverse target domains becomes necessary. Despite the recent extensive attention given to zero-shot image anomaly detection~\cite{arc,winclip,gu2024filo,cao2024adaclip,zhou2024anomalyclip}, the zero-shot setting in the complex surveillance video domain remains under-explored~\cite{aich2023cross}.

 % \textbf{Bottom:} Skeleton-based methods still face challenges under the ZS-VAD setting, including difficulty detecting indiscernible anomalies and misclassifying unseen normal events.

% \begin{figure}[t]
%     \centering
%     % 第一行
%     \begin{subfigure}[b]{0.45\textwidth}
%         \centering
%         \includegraphics[width=\textwidth]{intro1_1.pdf} 
%         \caption{The ZS-VAD potential of skeleton-based approach.}
%         \label{fig:sub1}
%     \end{subfigure}
%     % \vspace{0.5cm} % 图之间的垂直间距
%     % 第二行
%     \begin{subfigure}[b]{0.44\textwidth}
%         \centering
%     \includegraphics[width=\textwidth]{intro1_2.pdf} 
%         \caption{The remaining limitations of prevalent skeleton-based methods under the ZS-VAD setting.}
%         \label{fig:sub2}
%     \end{subfigure}
%     \caption{The illustration of our motivation.}
%     \label{fig1}
% \end{figure}

The challenges of ZS-VAD come from significant variations in visual appearance and human activities across different video domains. While frame/object-based methods~\cite{hf2,jigsaw,MULDE} have been prominent in video anomaly detection, their performance will degrade when adapting to new scenes due to visual feature distribution shifts. In another view, skeleton-based methods~\cite{morais,moprl,stgnf,MoCoDAD} utilize mature pose detection systems~\cite{cao2019openpose,fang2022alphapose} to obtain skeleton data, learn to encode features via self-supervision tasks~\cite{MoCoDAD,moprl}, and then calculate the anomaly score. They are effective for identifying human behavior anomalies, which are popular in the VAD task due to their superior efficiency and performance. skeleton-based methods also have inherent generalizable advantages
in achieving ZS-VAD as it eliminates domain disparities both in
background and human appearance.

% {As shown in Fig.~\ref{intro2}, we also observe that skeleton-based methods demonstrate better performance than frame/object-based methods on the unseen target domain, likely due to their inherent generalizable advantages in eliminating domain gaps in human appearance and scene background. }

% This suggests a promising approach to ZS-VAD.

\begin{figure}[t]
\begin{center}
\includegraphics[scale=0.7]{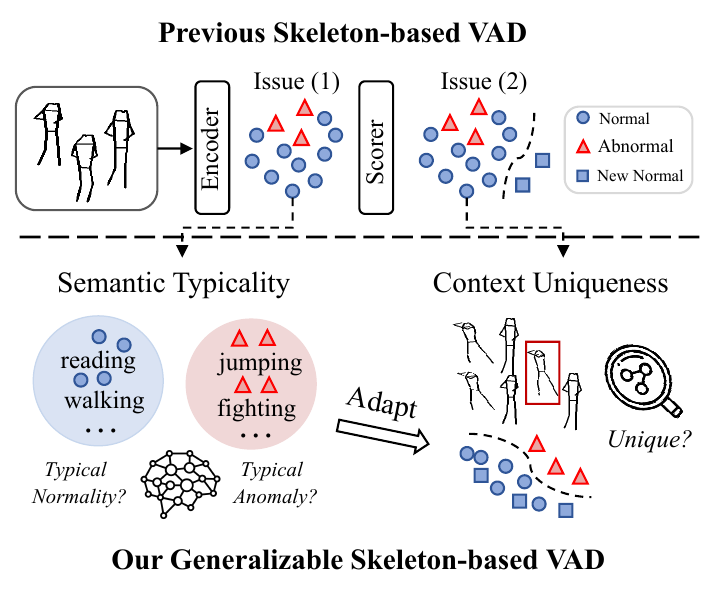}
\end{center}
\caption{{An illustration of skeleton-based VAD paradigm comparison. Previous approaches suffer from two main issues: \textbf{(1)} low-level representations and \textbf{(2)} domain-limited normal boundary. Our method enhances generalizability via action semantic typicality learning and context uniqueness analysis.} }\label{intro1}
\end{figure}

However, as shown in Fig.~\ref{intro1}, existing skeleton-based VAD methods still suffer from several limitations: \textbf{(1) Low-level skeleton representations.} They learn normal distribution of skeleton patterns using self-supervised tasks, such as skeleton prediction~\cite{MoCoDAD}, reconstruction~\cite{moprl}, or coordinate-based normalizing flows~\cite{stgnf}. Without semantic supervision signals, such methods fail to capture higher-level action patterns, making them unable to distinguish novel anomaly patterns similar to normal patterns and sensitive to noise. \textbf{(2) Domain-limited normality boundary.} They blindly rely on training-data-defined normality boundaries, leading to the misclassification of unseen normal events as anomalies. Both limitations hinder their generalization to unseen scenes with varying normal and abnormal patterns. This leads to a question: ``\textit{Can we further unlock the potential of skeleton in ZS-VAD with generalizable representation learning and prior injection?}''

% Within this line of work, ZS-VAD seems to become an ill-posed problem, because the new scene’s decision boundary is unattainable.

% \begin{figure}[t]
% \begin{center}
% \includegraphics[scale=0.54]{intro1_s.pdf}
% \end{center}
% \caption{{Skeleton-based methods~\cite{MoCoDAD,stgnf} demonstrate better performance in the target domain, where the models are trained on ShanghaiTech~\cite{shanghaitech} (Source) and tested on UBnormal~\cite{acsintoae2022ubnormal} (Target).} }\label{intro2}
% \end{figure}
% \begin{figure}[t]
% \begin{center}
% \includegraphics[scale=0.55]{intro1.pdf}
% \end{center}
% \caption{(a) The illustration of zero-shot video anomaly detection.
% (b) Performance on four benchmarks~\cite{shanghaitech,acsintoae2022ubnormal,nwpu,ucf}}\label{fig1}
% \end{figure}

% Although recently~\cite{zanella2024harnessing} proposes to leverage
% large visual language models to tackle the ZS-VAD task, it relies on multi-stage reasoning and the coordination of multiple large models, posing challenges for widespread
% deployment. 

To address this question, we reflect on how human observers judge normal and abnormal behavior in a new scenario. As shown in Fig.~\ref{intro1}, we first identify the types of individual actions in the video and consider whether they are normal or abnormal based on our experiential knowledge of normality and abnormality, which is referred to as typicality. For instance, a pedestrian walking would be considered normal, while a fight or scuffle would be deemed abnormal. Secondly, for atypical normal or abnormal scenarios, we integrate the behaviors of all individuals in the video to observe if any individual's behavior significantly differs from others, as anomalies are usually rare and unique, referred to as context uniqueness.

% Previous methods depend on domain-specific data to obtain a normality boundary, limiting their adaptability to new scenes that have different normal and abnormal behavior patterns.

Based on these complementary priors, we propose a novel skeleton-based zero-shot video anomaly detection framework, which captures both typical anomalies guided by language prior and unique anomalies in spatio-temporal contexts. First, we introduce a language-guided typicality modeling module to achieve high-level semantic understanding beyond previous low-level representations. Specifically, it projects skeleton snippets into language-aligned action semantic space and distills LLM’s knowledge of typical normal and abnormal behaviors during training. Secondly, to derive scene-adaptive boundaries, we propose a context uniqueness analysis module at test time. It finely analyzes
the spatio-temporal differences between skeleton snippets to get an adaptive understanding of target scene activities. Without using any training samples from the target domain, we achieve state-of-the-art results on four large-scale VAD datasets: ShanghaiTech~\cite{shanghaitech}, UBnormal~\cite{acsintoae2022ubnormal}, NWPU~\cite{nwpu}, UCF-Crime~\cite{ucf}, featuring over 100 unseen
surveillance scenes. Our contributions are as follows:
\begin{itemize}
\item We propose a skeleton-based video anomaly detection framework that learns action typicality and uniqueness, enabling generalization across diverse target scenes.
\item We propose a language-guided typicality modeling module that projects skeleton snippets into a generalizable semantic space and distills LLM's knowledge of typical normal and abnormal behaviors during training. 
\item We propose a test-time uniqueness analysis module to finely analyze the spatio-temporal differences between skeleton snippets and derive scene-adaptive boundaries between normal and abnormal behavior.
\end{itemize}

% effectively learns the typical distribution of normal and abnormal behavior based on LLM's knowledge.
The rest of this paper is organized as follows. We review the related work in Section \ref{related}. Section \ref{method} describes the technical
details of our proposed method. Section \ref{exper} presents the
experiment details and results. Finally, we summarize the
paper in Section \ref{conclusion}.

\section{Related Work}\label{related}
\textbf{Video anomaly detection. }Most previous video anomaly detection studies can be grouped into frame-based~\cite{shanghaitech,hf2,ucf}, object-centric~\cite{jigsaw,semantic,MULDE}, and skeleton-based methods~\cite{gepc,MoCoDAD,stgnf}. In this work, we focus on the skeleton-based methods, which detect anomalies in human activity based on preprocessed skeleton/pose data. Morais et al.~\cite{morais} propose an anomaly detection method that uses an RNN network to learn the representation of pose snippets, with prediction errors serving as anomaly scores. GEPC~\cite{gepc} utilizes autoencoders to learn pose graph embeddings, generates soft assignments through clustering, and uses a Dirichlet process mixture to determine anomaly scores. To model normal diversity, MoCoDAD~\cite{MoCoDAD} leverages diffusion probabilistic models to generate multimodal future human poses. FG-Diff~\cite{fg-diff}    guides the diffusion model with observed
 high-frequency information and prioritizes the reconstruction of
 low-frequency components, enabling more accurate and robust
 anomaly detection. STG-NF~\cite{stgnf} proposes a simple yet effective method by establishing normalized flow~\cite{kingma2018glow} from normal pose snippets to obtain normal boundaries. DA-Flow~\cite{daflow} proposes a lightweight dual attention module for capturing cross-dimension interaction relationships in spatio-temporal skeletal
data.  However, these methods rely on training with normal data from the target domain, while overlooking the semantic understanding of human behavior, which makes it difficult to ensure performance in scenarios where the target data is unavailable. 

\textbf{Zero-shot anomaly detection. }
Thanks to the development of vision-language models, zero-shot anomaly detection has received a lot of attention~\cite{winclip,batch,gu2024filo,aota2023zero,arc,VAND,zhou2024anomalyclip}, especially in the field of image anomaly detection~\cite{survey}. The pioneering work is WinCLIP~\cite{winclip}, which utilizes CLIP~\cite{clip}'s image-text matching capability to distinguish between unseen normal and abnormal anomalies. Building on that, AnomalyCLIP~\cite{zhou2024anomalyclip} proposes to learn object-agnostic text prompts that capture generic normal and abnormal patterns in an image. AdaCLIP~\cite{cao2024adaclip}  introduces two types of learnable prompts to enhance CLIP's generalization ability for anomaly detection. Despite the success in the image domain, only a few works~\cite{aich2023cross,guo2024ada} have ventured into zero-shot video anomaly detection with underwhelming performance. Although recently ~\cite{zanella2024harnessing}  proposes to leverage large visual language models for zero-shot video anomaly detection, it requires multi-stage reasoning and the collaboration of multiple large models, making it less user-friendly. We aim to develop a lightweight, user-friendly, and easily deployable zero-shot anomaly detector starting from skeleton data. Our work shares some similarities with a recent study~\cite{sato2023prompt}. However, we emphasize that our approach differs significantly from~\cite{sato2023prompt} in the following ways:~\textbf{1) Different tasks:} It addresses abnormal action recognition, involving no more than two individuals in a short video, while ours requires temporally localizing abnormal events in real surveillance videos.  \textbf{2) Novel perspective:} We combine the action typicality and uniqueness priors to address zero-shot anomaly detection challenges in video surveillance scenes.

\section{Method}\label{method}
\subsection{Overview}
The objective of ZS-VAD is to train one model that can generalize to diverse target domains. Formally, let $\mathcal{V}^{train}$ be a training set from source video domain and $\{\mathcal{W}^{test}_1, \mathcal{W}^{test}_2, ..., \mathcal{W}^{test}_N\}$ be multiple test sets from target video domain. The test videos are annotated at the frame level with labels ${l_i}\in\{0,1\}$, and the model is required to predict frame-level anomaly score.
In this work, we focus on the skeleton-based paradigm, as it is computation-friendly and can benefits ZS-VAD by reducing the domain gap in both background and appearance. 

\begin{figure*}[t]
\begin{center}
\includegraphics[scale=0.385]{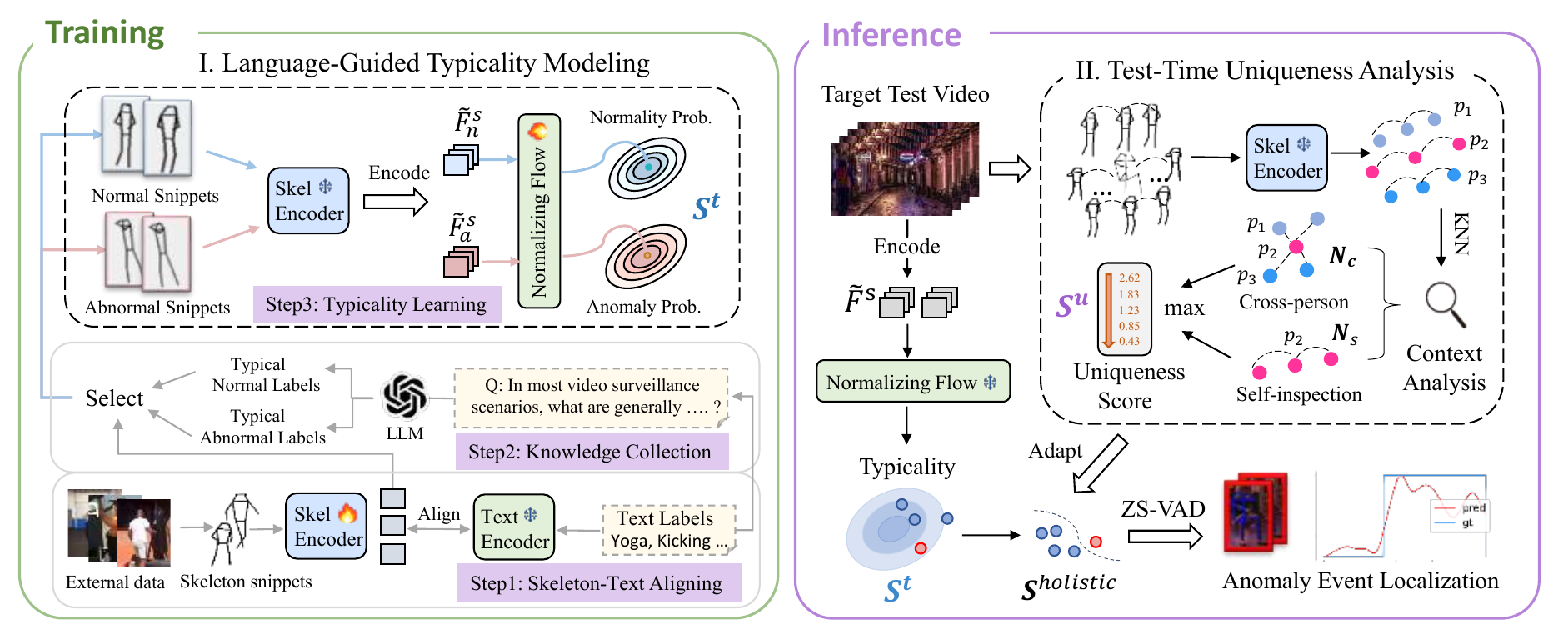}
\end{center}
\caption{{Overview of our approach for skeleton-based zero-shot video anomaly detection.~\textbf{I.} Language-guided typicality modeling in the training phase. It projects skeleton snippets into the action semantic space, collects typicality knowledge from LLM, and then effectively learns the typical distribution of normal and abnormal
behavior. (\textbf{Only the black dashed boxes are used during inference.}) ~\textbf{II.} Test-time uniqueness analysis in the inference phase. It finely analyzes the spatio-temporal differences between skeleton snippets and derives scene-adaptive boundaries between normal and abnormal behavior.}} \label{fig2}
\end{figure*}

Fig. \ref{fig2} overviews our proposed approach. Our model tackles the ZS-VAD problem from the perspective of action typicality and uniqueness learning. Firstly, to obtain a high-level semantic understanding, we propose a Language-Guided Typicality Modeling module that projects skeleton snippets into action semantic space and distills LLM’s knowledge of typical normal and abnormal behaviors during training. Secondly, to get scene-adaptive decision boundaries, we propose a Test-Time Uniqueness Analysis module that finely
analyzes the spatio-temporal differences between skeleton snippets. During inference on unseen VAD datasets, our model integrates typicality scores and uniqueness scores of human behavior to provide a holistic understanding of anomalies.

\subsection{Language-guided Typical Modeling}\label{sec1}
Unlike previous works that learn low-level skeleton representations via self-supervised tasks~\cite{moprl,stgnf,MoCoDAD}, this module aims to obtain a high-level semantic understanding of human behavior. It learns language-aligned action features and scene-generic distributions of typical distribution with distillation of LLM's knowledge during training. Specifically, this module consists of skeleton-text alignment, typicality knowledge selection, and typicality distribution learning. During inference, it can predict typicality anomaly scores with only a lightweight skeleton encoder and a normalizing flow module. 

\textbf{Skeleton-text alignment.} For achieving a generalizable semantic understanding of human behavior, we first propose to align the skeleton snippets with the corresponding semantic labels. For such skeleton-text pairs, we utilize external action recognition datasets (e.g., Kinect-400~\cite{kinect} as the training set instead of specific VAD datasets (e.g., ShanghaiTech). The raw skeleton data of an action video is typically formally represented as $\mathbf{X}_i\in \mathbb{R}^{C\times J\times L\times M}$, where $C$ is the coordinate number, $J$ is the joint number, $L$ is the sequence length, and $M$ is the pre-defined maximum number of persons, respectively. In addition, each video is annotated with a text label $g_i$ representing the action class, which can also be transformed into a one-hot class vector $\mathbf{y}_i$.

Compared to action recognition tasks~\cite{stgcn,liu2023parallel,wang2024learning} that only predict video-level categories, the VAD task is more fine-grained, focusing on frame-level anomaly scores.
Therefore, we decompose the original sequences into multiple short skeleton snippets $\mathbf{A}_i \in \mathbb{R}^{C\times J\times T}$ using a sliding window, and discard snippets that are composed of zeros, where $T$ is the length of a snippet. For the snippets from the same action video, they share the same labels and undergo a normalization operation like STG-NF~\cite{stgnf} to make different snippets independent.
 Inspired by the recent multimodal alignment works~\cite{wang2021actionclip,gap}, we then perform a skeleton-text alignment pretraining procedure to learn the discriminative representation. The procedure is built with a skeleton encoder $E^s$ and a text encoder $E^t$, for generating skeleton features $\mathbf{F}^s$ and text features $\mathbf{F}^t$, respectively. Additionally, the skeleton encoder also predicts a probability vector $\hat{\mathbf{y}}_i$ using a fully-connected layer. The training loss consists of a KL divergence loss and a cross-entropy classification loss following GAP~\cite{gap}. This skeleton-text alignment procedure effectively guides the projection of skeleton snippets into language-aligned action semantic space beyond previous VAD works~\cite{moprl,stgnf,MoCoDAD} that learns low-level skeleton representation.
%  which are presented as follows:
%  % \small{
% \begin{equation}
% \begin{aligned}
%        \mathcal{L}_{s} = \frac{1}{2N}\sum_{i=1}^N
%     &\left(\mathcal{L}_k (p^{s2t}(\mathbf{F}^s_i), y^{s2t}_i)+\mathcal{L}_k (p^{t2s}(\mathbf{F}^t_i), y^{t2s}_i)\right)  \\ 
%     &+\frac{1}{N}\sum_{i=1}^N\mathcal{L}_{cls}(\hat{\mathbf{y}},\mathbf{y}),
% \end{aligned}
% \end{equation}
% % }
% where $\mathcal{L}_k$ denotes KL divergence loss, and the $N$ is the sample count.

% $y^{s2t}$ and $y^{t2s}$ denote the ground-truth matching labels, and Based on the loss function in CLIP~\cite{clip}, the networks are optimized by contrasting skeleton-text pairs in two directions within the batch:
% \begin{equation}
% \begin{aligned}
%  p^ {s2t} (\mathbf{F}^s_i)=  \frac {\mathrm{exp}(\mathrm{sim}(\mathbf{F}^s_i,\mathbf{F}^t_i)/\tau )}{\sum _ {j=1}^ {B}\mathrm{exp}(\mathrm{sim}(\mathbf{F}^{s}_i,\mathbf{F}^t_j)/\tau )} , \quad
% p^ {s2t} (\mathbf{F}^t_i)=  \frac {\mathrm{exp}(\mathrm{sim}(\mathbf{F}^t_i,\mathbf{F}^s_i)/\tau )}{\sum _ {j=1}^ {B}\mathrm{exp}(\mathrm{sim}(\mathbf{F}^{t}_i,\mathbf{F}^s_j)/\tau )} ,
% \end{aligned}
% \end{equation}
% where $B$ is the batch size, and $\mathrm{sim}(\cdot)$ denotes the cosine similarity.

\textbf{Typicality knowledge selection. } In most video surveillance scenarios, some behaviors are generally considered normal or abnormal, which constitute a scene-generic set. Therefore, training a typicality-aware capability is one of the promising ways to achieve ZS-VAD. 

% Thanks to the cutting-edge advancements of Large Language Models (LLMs), we propose to distill their prior knowledge about generic normality and abnormality during training. 
{Given that we adopt the Kinect-400 dataset as our external auxiliary training set, we aim to identify typical normal and abnormal action categories from the 400 classes and then use the corresponding videos to train our generalized anomaly detector. In this phase, the LLM acts as an automated offline knowledge extractor that pinpoints the typical normal and abnormal actions consistent with human experiential knowledge.}
In detail, we give the LLM a prompt $\mathcal{P}$: ``\textit{In most video surveillance scenarios, what are generally considered as normal actions and abnormal actions among these actions (Please identify the 20 most typical normal actions and 20 most typical abnormal actions, ranked in order of decreasing typicality). The action list is \textless$\mathcal{T}$\textgreater}'', where $\mathcal{T}$ refers to the set of all 400 action class labels. {We then filter out any actions generated by the model that are not present in $\mathcal{T}$.} The selected typicality labels are shown in Table~\ref{labels}.
% Based on the pre-trained skeleton-text representation, we aim to use a LLM as our knowledge engine to collect typical normal and abnormal data from the massive skeleton snippets.
 \begin{table}[t]
\setlength\tabcolsep{3pt}
\small
\centering
\begin{tabular}{cc}
   \toprule
Type & Typicality action list\\
\midrule
\multirow{2}*{Normal} & reading book, reading newspaper, walking the dog, \\
&washing dishes, washing feet, washing hair...\\
\midrule
\multirow{2}*{Abnormal}& skydiving, scuba diving,
ski jumping, sailing, \\
&rock climbing, snowboarding, ice skating...\\
   \bottomrule
\end{tabular}
\caption{The generated typicality labels.} \label{labels}
\end{table}
The large language model will respond with a list of typical normal action classes  $\mathcal{T}^n$ and a list of typical abnormal action classes  $\mathcal{T}^a$, which can be formalized as:
\begin{equation}
   \mathcal{T}^n, \mathcal{T}^a = \mathrm{O}_{LLM}(\mathcal{P}, \mathcal{T}),
\end{equation}
where $\mathcal{T}^n$ and $\mathcal{T}^a$ are the subsets of $\mathcal{T}$, and $\mathcal{O}_{LLM}$ denotes the offline LLM used for initial typicality label generation. Note that the LLM is only needed to \textbf{be used once during training} for auxiliary data selection, while inference is not. 

% Therefore, we use the powerful LLM, ChatGPT~\cite{openai2022chatgpt}, as our offline knowledge engine, 

After knowing the action categories of typicality, we first collect the data of these selected categories and then proceed to select the high-quality snippets from them. This is because 1)
Some snippets contain noise, such as errors in pose detection and tracking. 2)
In an abnormal action sequence, not all the snippets are abnormal. Therefore, we use the skeleton-text similarity score to select the high-quality skeleton snippets, which is formulated as:
\begin{equation}
   {\mathcal{M}}^{x}=\{\mathrm{arg}\,  \mathop{\mathrm{top {\beta}}^x}_i\left(\mathrm{sim}\left(\mathbf{F}^s_i,\mathbf{F}^t_{g_i}\right)\right)\, | \, g_i\in\mathcal{T}^{x}\},
\end{equation}
where $\mathcal{M}^x$ refers to the selected snippets index, $g_i$ denotes the text label of snippet $i$, and $\beta$ denotes the selection ratio. The superscript $x$ represents $n$ or $a$, indicating normal and abnormal, respectively. Using the  index $\mathcal{M}^x$, we obtain the corresponding skeleton data $\tilde{\mathbf{A}}^n$ and $\tilde{\mathbf{A}}^a$, as well as skeleton features $\tilde{\mathbf{F}}_n^s$ and $\tilde{\mathbf{F}}_a^s$.

\textbf{Typicality distribution learning. } As shown in Fig.~\ref{fig2}, after obtaining the data, we proceed to model the feature distribution of typical behavior. Normalizing Flow (NF)~\cite{kingma2018glow} provides a robust framework for modeling feature distributions, transforming this distribution through a series of invertible and differentiable operations. Consider a random variable $\mathbf{X}\in \mathbb{R}^D$ with target distribution $p_X(x)$, and a random variable Z follows a spherical multivariate Gaussian distribution. A bijective map $f: X\leftrightarrow Z$ is then introduced, which is composed of a sequence of transformations: $f_1\circ f_2\circ...\circ f_K$. According to the variable substitution formula, the log-likelihood of X can be expressed as:
\begin{equation}
    \log p_{X}(x)=\log p_{Z}(f(x))+\sum_{i=1}^{K}\log \left|\det(\frac{df_{i}}{df_{i-1}})\right|.
\end{equation}
Using such a transformation, the feature distribution of typicality behavior is effectively modeled. Specifically, the bijective maps for the normal features and abnormal features are $f: X_n \leftrightarrow Z_n$ and $f: X_a \leftrightarrow Z_a$, respectively. Here, the log-likelihood of $Z_n$ and $Z_a$ are as follows:
\begin{equation}
\begin{aligned}
&\log p_{Z_n}(z)=\mathrm{Con}-\frac{1}{2}(z-\mu_n)^{2},
\\
&\log p_{Z_a}(z)=\mathrm{Con}-\frac{1}{2}(z-\mu_a)^{2},
\end{aligned}
\end{equation}
where $\mathrm{Con}$ is a constant, and $u_n$ and {$u_a$} are the centers of the Gaussian distributions ({$|u_n-u_a|\gg 0$}), respectively. During training, the normalizing flow is optimized to increase the log-likelihood of the skeleton features $\mathbf{F}^s$ with the following loss:
\begin{equation}
    \mathcal{L}_n = -\log p_{X_n}(\mathbf{F}_n^s)-\log p_{X_a}(\mathbf{F}_a^s),
\end{equation}
During inference, the testing skeleton snippet $\mathbf{F}^s_i$ will be sent to the trained normalizing flow, outputting the typicality anomaly score as follows:
\begin{equation}
    \mathbf{S}^t_i = -\log p_{X_n}(\mathbf{F}^s_i), 
\end{equation}
where the normal skeletons will exhibit low $\mathbf{S}^t_i$, while the anomalies will exhibit higher $\mathbf{S}^t_i$. Our approach differs significantly from STG-NF~\cite{stgnf}. It takes low-level skeleton coordinates as inputs and only learns implicit spatio-temporal features, which struggle to generalize to new datasets without the normality reference of training data from the target dataset. Differently, we use action semantics as a generalizable representation for normalizing flow input and leverage experiential typicality labels to learn domain-general boundaries between normal
and abnormal behavior.

\subsection{Test-time Uniqueness Analysis}\label{sec2}
The goal of this component is to serve as a complementary perspective of typicality, deriving scene-adaptive boundaries by considering the context of the target scene. To this end, we propose a context uniqueness analysis module during the inference of the unseen VAD dataset.

Unlike action recognition datasets, surveillance videos contain richer contextual information, featuring longer temporal spans, larger numbers of people, and more diverse behavioral patterns. For such a video, $H$ skeleton sequences $\{\mathbf{X}_1, ..., \mathbf{X}_H\}$ are extracted, where each sequence comprises ${L_i}$-frame poses, represented as  $\mathbf{X}_i = \{\mathbf{P}_1, ..., \mathbf{P}_{L_i}\}$. Here, $\mathbf{P}_t\in \mathbb{R}^{J\times2}$ comprises $J$ keypoints, each defined by a pair of coordinate values. Targeted at frame-level anomaly scoring, the sequences are segmented into shorter skeleton snippets, denoted as $\mathbf{A}_i \in \mathbb{R}^{C \times J \times T}$, each of which is then individually scored based on its contextual information.

\textbf{Spatio-temporal context.} 
As shown in Fig.~\ref{fig2}, to gain a fine-grained context understanding of the scene, we construct \textit{two types of spatio-temporal context graphs}: a \textbf{cross-person graph} $\mathcal{G}^{c}$ and a \textbf{self-inspection graph} $\mathcal{G}^{s}$. The first graph is constructed by retrieving the \textit{feature nearest neighbors} among the surrounding skeleton snippets, while the second one is constructed by retrieving the \textit{feature nearest neighbors} from different time segments of the current person. In this way, we can \textbf{filter out some unrelated activities} and focus solely on behaviors related to the current individual.
Given a skeleton snippet $\mathbf{A}_i$ with feature $\mathbf{F}^s_i$, the cross-person graph is defined as $\mathcal{G}^{c}_i=\{\mathcal{V}^{c}_i,\mathcal{E}^{c}_i\}$, where $\mathcal{V}^{c}_i=\{\mathbf{A}_i,\mathcal{N}_{c}(\mathbf{A}_i)\}$ denotes  the node set and $\mathcal{E}_{c}^i=\left\{(i,j)\,|\,j \in \mathcal{N}_{c} \right\}$ denotes the edge set. Besides, during the preprocessing of skeleton snippets, $\mathbf{A}_i$ is associated with a human trajectory index $p_i$ and timestamp $t_i$. The neighborhood $\mathcal{N}_{c}$ is formulated as:
\begin{equation}
    \mathcal{N}_{c} = \left\{\mathbf{A}_j \,|\, d\left(\mathbf{F}_i^s, \mathbf{F}_j^s\right) \leq D^k_c, \,p_j\neq p_i \right\},
\end{equation}
where $d(\cdot)$ represents the Euclidean distance, and $D^k_c$ refers to the k-th smallest value for the \textbf{cross-person distances}. The second graph, which depicts self-inspection, is defined as $\mathcal{G}^{s}_i=\{\mathcal{V}^{s}_i,\mathcal{E}^{s}_i\}$, where $\mathcal{V}^{s}_i=\{\mathbf{A}_i,\mathcal{N}_{s}(\mathbf{A}_i)\}$ denotes the node set and $\mathcal{E}_{s}^i=\{(i,j)\,|\,j \in \mathcal{N}_{s} )$ denotes the edge set. Then, the neighborhood $\mathcal{N}_{s}$ is formulated as:
\begin{equation}
\begin{aligned}
     \mathcal{N}_{s} = \{\mathbf{A}_j \,|\, d\left(\mathbf{F}_i^s, \mathbf{F}_j^s\right) \leq D^k_s,  
     \,p_j = p_i,\, |t_i - t_j| > \alpha T  \},
\end{aligned}
\end{equation}
 where $D^k_s$ refers to the k-th smallest value for the \textbf{self-inspection distances}. $\alpha$ is a threshold that masks out the time before and after the current time window, as the individual's state tends to remain stable during adjacent periods.

\textbf{Uniqueness scores.} Since abnormal activities are rare, anomalies in real-world surveillance videos often differ from other activities in both spatial and temporal context, which is referred to as uniqueness. Based on the pre-trained discriminative skeleton features, uniqueness can be represented as the feature distances between a query node and other nodes in the built graph. Specifically, the uniqueness score $\mathbf{S}^u$ for individual $i$ is obtained by taking the larger one of the cross-person and self-inspection distances, formulated as follows:
\begin{equation}
    \mathbf{S}^u_i = \mathrm{max} \, \{ \sum\limits_{j\in\mathcal{N}_{c}\left(\mathbf{A}_i\right)} d\left(\mathbf{F}_i^s, \mathbf{F}_j^s\right), \sum\limits_{j\in\mathcal{N}_{s}\left(\mathbf{A}_i\right)} d\left(\mathbf{F}_i^s, \mathbf{F}_j^s\right) \,\,\}.
\end{equation}
\textbf{Holistic anomaly scoring.} 
By integrating the complementary typicality $\mathbf{S}^t_i$ scores and the uniqueness scores $\mathbf{S}^u_i$, our model can capture both typical anomalies in language prior and unique anomalies in spatio-temporal contexts. This helps gain a comprehensive understanding of anomalies in new scenes, where the holistic anomaly score of individual $i$ is defined as:
\begin{equation}
    \mathbf{S}_i = \frac{\mathbf{S}^t_i - \mathbf{S}^t_{mean}}{\mathbf{S}^t_{std}} + \frac{\mathbf{S}^u_i - \mathbf{S}^u_{mean}}{\mathbf{S}^u_{std}}.
\end{equation}
Finally, the frame-level anomaly scores are obtained by taking the highest score among all individuals within each frame. If any individual is considered anomalous, the entire frame is classified as anomalous. For frames where no individuals are detected, it is classified as a normal frame. In this condition, the anomaly score is assigned the minimum value among all scores in that video, following the approach in~\cite{stgnf}.

%  \begin{table}[t]
% \setlength\tabcolsep{3.8pt}
% \footnotesize
% \centering
% \begin{tabular}{ccccc}
%    \toprule
%  &\multirow{2}*{Year}&\multirow{2}*{Resolution}&Test Video&Scenes\\
%  &&&Num.&Num.\\
%   \midrule
% ShanghaiTech~\cite{shanghaitech}&2018&480$\times$856&107&13\\
% UBnormal~\cite{acsintoae2022ubnormal}&2022&720$\times$1280&211&29\\
%   NWPU~\cite{nwpu}&2023&multiple&242&43\\
%  UCF-Crime~\cite{ucf}&2018&240$\times$320&290&\textgreater50\\
%    \bottomrule
% \end{tabular}
% \caption{The details of the zero-shot benchmarks. } \label{dataset}
% \end{table}
 
\section{Experiments}\label{exper}
\subsection{Dataset and Implementation Details}
\textbf{Dataset.} 
The training of our model is conducted on the Kinect-400-skeleton dataset~\cite{kinect,stgcn}, while the ZS-VAD capability of our model is evaluated on four large-scale VAD datasets: ShanghaiTech~\cite {shanghaitech}, UBnormal~\cite{acsintoae2022ubnormal}, NWPU~\cite{nwpu} and UCF-Crime~\cite{ucf}. Note that we only use the test set of these four VAD datasets. 

\begin{table}[t]
\setlength\tabcolsep{5pt}
\small
\centering
\begin{tabular}{cccccc}
   \toprule
 &\multirow{2}*{Resolution}&Test Video&Scenes&Snippet \\
 &&Num.&Num.&Num.\\
  \midrule
% Kinect-400~\cite{stgcn,kinect}&multiple&-&-& 3,490,558\\
ShanghaiTech~\cite{shanghaitech}&480$\times$856&107&13&156,571\\
UBnormal~\cite{acsintoae2022ubnormal}&720$\times$1280&211&29&315,416\\
  NWPU~\cite{nwpu}&multiple&242&43&723,490\\
 UCF-Crime~\cite{ucf}&240$\times$320&290&\textgreater50&152,231*\\
   \bottomrule
\end{tabular}
\caption{The details of our zero-shot video anomaly detection benchmarks. Each snippet contains 16 frames of skeleton data with a 1-frame interval, while the snippets of UCF-Crime* are sampled with a 16-frame interval as its videos are too long.} \label{dataset}
\end{table}
% Table \ref{dataset} summarizes the dataset characteristics.
% Table~\ref{tab1} depicts the combination of datasets used during the training and testing phases.
For the training of our model, we use the external Kinect-400-skeleton~\cite{stgcn} dataset. It is not intended for VAD tasks but for action recognition, which is gathered from YouTube videos covering 400 action classes. We utilize the preprocessed skeleton data obtained from ST-GCN~\cite{stgcn} for training.
% The dataset includes 250,000 training and 19,000 validation video clips, each lasting 10 seconds and recorded at 30 frames per second. 
% 2) VAD datasets for evaluation: \textit{ShanghaiTech} is a widely-used benchmark for one-class video anomaly detection, which consists of 107 test videos from 13 different scenes. \textit{UBnormal} is
% \textbf{(2)}  VAD datasets for evaluation, with a total of over 100 unseen surveillance scenes: \textit{ShanghaiTech} contains  107 test videos from 13 different scenes. \textit{UBnormal} includes 211 test videos from 29 different scenes. \textit{NWPU} is a newly published dataset containing 242 testing videos from 43 scenes. \textit{UCF-Crime} includes 290 testing videos from more than 50 scenes. 
For evaluation, we take four VAD-relevant datasets. Compared to some early VAD benchmarks~\cite{avenue, ped} that involve single scenes staged and captured at one location, the four datasets we evaluated are more extensive, encompassing a wider variety of scenes. Consequently, these four datasets are better suited for testing the model's zero-shot capabilities and assessing its cross-scenario performance. 
The details are summarized in Table~\ref{dataset} and the following descriptions. 
\textbf{(1)} ShanghaiTech (SHT). It is a widely-used benchmark for one-class video anomaly detection, which consists of 330 training videos and 107 test videos from 13 different scenes. 
\textbf{(2)} UBnormal (UB).  It is a synthetic dataset with virtual objects and real-world environments. It consists of 186 training videos and 211 test videos from 29 different scenes. 
\textbf{(3)} NWPU. It is a newly published dataset that contains some scene-dependent anomaly types. It comprises 305 training videos and 242 testing videos from 43 scenes.
\textbf{(4)} UCF-Crime (UCF-C).  It is a large-scale dataset with 1900 long untrimmed surveillance videos. The 290 testing videos are used for our evaluation.

 % \begin{wraptable}{r}{5.6cm}
  \begin{table}[t]
\setlength\tabcolsep{4.0pt}
% \footnotesize
\small
\centering
\begin{tabular}{c|c|c|cccc}
   \toprule
\multirow{2}*{Paradigm} & \multirow{2}*{Method} & Training & \multicolumn{4}{c}{Testing Set}\\
  &&VAD&SHT$_{13}$&UB$_{29}$&NWPU$_{43}$&UCF-C$_{> 50}$\\
  \midrule
   % \multicolumn{2}{l}{\textit{\quad\quad LVLM}}&&&&\\
\multirow{2}*{LVLM} &Imagebind~\cite{girdhar2023imagebind}&\ding{56}&-&-&-&\textcolor{gray}{55.8}\\
% LLaVA-1.5~\cite{liu2023improvedllava}&\ding{56}&-&-&-&\textcolor{gray}{72.8}\\
&LAVAD~\cite{zanella2024harnessing}&\ding{56}&-&-&-&\textcolor{gray}{80.3}\\
\midrule
  % \multicolumn{2}{l}{\textit{\,\,\,Frame/Object}}&&&&\\
\multirow{2}*{Frame/Object}&HF2-VAD~\cite{hf2}&SHT& \textcolor{gray}{76.2}& 59.5& 58.3&52.9 \\
&Jigsaw-VAD~\cite{jigsaw}&SHT& \textcolor{gray}{84.3}&58.6&61.1&53.3\\
% object&MULDE~\cite{MULDE}&SHT& & & & \\
% object&MULDE~\cite{MULDE}&UB& & & & \\
\midrule
% skeleton&MoPRL~\cite{moprl}&SHT& & & & \\
% \multicolumn{2}{l}{\textit{\quad\,\, Skeleton}}&&&&\\
\multirow{5}*{Skeleton} &MocoDAD~\cite{MoCoDAD}&SHT&\textcolor{gray}{77.6}& 67.0& 56.4& 51.8\\
&MocoDAD~\cite{MoCoDAD}&UB& 76.0& \textcolor{gray}{68.4}&56.6 &52.0 \\
&STG-NF~\cite{stgnf}&SHT& \textcolor{gray}{85.9}&68.8 & 57.6& 51.6\\
&STG-NF~\cite{stgnf}&UB&83.0 &\textcolor{gray}{71.8} &57.9 &51.9\\
% \rowcolor{blue!8}
&\bf{Ours}&\ding{56}&\bf{84.1}&\bf{74.5}&\bf{62.1}&\bf{62.7}\\
   \bottomrule
\end{tabular}
\caption{\textbf{Zero-shot} video anomaly detection performance on the four large-scale datasets, ShanghaiTech, UBnormal, NWPU, and UCF-Crime, where the subscript denotes the number of scenes. The \textcolor{gray}{gray} text indicates the full-shot performance.} \label{tab1}
\end{table} 

\begin{table}[t]
\setlength\tabcolsep{8pt}
\small
\centering
\begin{tabular}{c|c|c|c|c|c}
   \toprule
  \multirow{2}*{Setting}&\multirow{2}*{Method}&Training&Testing&Training&Testing\\
  &&on SHT&on SHT&on UB&on UB\\
  \midrule
zero-shot&Ours&\ding{56}&84.1&\ding{56}&\underline{74.5}\\
\midrule
  \multirow{7}*{full-shot}&HF2-VAD~\cite{hf2}&\ding{51}&76.2&\ding{51}&-\\
   &Jigsaw-VAD~\cite{jigsaw}&\ding{51}&84.3&\ding{51}&-\\
&SSMTL++~\cite{barbalau2023ssmtl++}&\ding{51}&83.8&\ding{51}&62.1\\
  &GEPC~\cite{gepc}&\ding{51}&76.1&\ding{51}&53.4\\
  &MocoDAD~\cite{MoCoDAD}&\ding{51}&77.6&\ding{51}&68.4\\
  &STG-NF~\cite{stgnf}&\ding{51}&\underline{85.9}&\ding{51}&71.8\\
  & FG-Diff~\cite{fg-diff}&\ding{51}&-&\ding{51}&68.9\\
  % \rowcolor{blue!8}
  &\textbf{Ours-full}&\ding{51}&\bf{86.0}&\ding{51}&\bf{78.2}\\
   \bottomrule
\end{tabular}
\caption{Our zero-shot performance vs. SOTA full-shot performance. We also provide a version named Ours-full to evaluate our method under the popular full-shot setting.} \label{full}
% \end{wraptable}
\end{table}

\textbf{Implementation Details.}
For a fair comparison, we directly use the skeleton data of ShanghaiTech and UBnormal from STG-NF~\cite{stgnf}. For NWPU and UCF-Crime, as they do not have open-source skeleton data, we resort to utilizing AlphaPose~\cite{fang2022alphapose} for data extraction. We use a segment window $T$ = 16 and a stride of 1 to divide each sequence into snippets. Specifically, we use a stride of 16 for UCF-Crime because its videos are too long. For the backbone, we use multi-scale CTR-GCN~\cite{ctr} (2.1M) as the skeleton encoder and use a 4-layer feature normalizing flow~\cite{BGAD} (2.9M) to model the normality probability. During training, we use the 
``ViT-B/32'' CLIP~\cite{clip} as the text encoder, and \texttt{GPT-3.5-Turbo} as our knowledge engine . During inference, these two models are removed. For the hyperparameters, the batch size is set to 1024, and the Adam optimizer is used with a learning rate of 0.0005. Additionally,  $\beta^n$, $\beta^a$, $k$, and $\alpha$ are set to 90\%, 10\%, 16, and 4, respectively.
For the evaluation metrics, we follow common practice ~\cite{shanghaitech,ucf,stgnf} by using the micro-average frame-level AUC as the evaluation metric, concatenating all frames and calculating the score.

% utilizing the area under the curve (AUC) of the frame-level receiver operating characteristics (ROC) This approach
% We use the CTR-GCN as the skeleton backbone, and use CLIP as the text encoder.
% Furthermore, we normalize each pose segment to have a zero mean and unit variance following the approach in~\cite{stgnf}.
% For the selection of typicality knowledge, we gather 80\% of normal data and 10\% of abnormal data. During training, t
% \textbf{Evaluation metrics.} For performance evaluation, we follow common practice ~\cite{shanghaitech,ucf,stgnf} by utilizing the area under the curve (AUC) of the frame-level receiver operating characteristics (ROC) as the evaluation metric. This approach involves concatenating all frames and calculating the score, also referred to as micro-averaged AUC~\cite{auc}.

\subsection{Main Results}
We conduct a comprehensive comparison of the performance of ZS-VAD, comparing the frame-based/object-based~\cite{hf2,jigsaw}, skeleton-based~\cite{MoCoDAD, stgnf}, and LVLM-based methods~\cite{girdhar2023imagebind,zanella2024harnessing}. 

\textbf{Comparison with frame/object-based methods.} 
 We use their open-source checkpoints trained on the ShanghaiTech to evaluate the zero-shot performance on the remaining three VAD datasets. As shown in Table~\ref{tab1}, their generalization capabilities on new scene datasets are relatively poor due to the influence of human appearance and background variations.
 
\textbf{Comparison with skeleton-based methods.}
 % The existing methods require VAD source data to train normal boundaries, and thus w
% Table \ref{tab1} shows the
% comparison results of ZS-VAD performance on the four VAD benchmarks. 
We use their open-source checkpoints trained on the ShanghaiTech or UBnormal to evaluate on the remaining three VAD datasets.
 The performance of prevalent skeleton-based methods is still underwhelming due to a lack of understanding of complex normal and abnormal behaviors without target training data.
Compared with our baseline STG-NF, our proposed method improves the frame-level AUC-ROC by 1.1\% on ShanghaiTech, 5.7\% on UBnormal, 4.2\%  on NWPU, and 10.8\% on UCF-Crime. We also compare their performance in the full-shot setting, where target domain data is used for training. Table~\ref{full} shows that our zero-shot approach can achieve comparable or even superior results to SoTA full-shot performance. To evaluate our method under the popular \textbf{full-shot} setting, we train our normalizing flow only on VAD normal data to model the normal distribution and test it on the same domain. The results outperform state-of-the-art (SOTA) full-shot methods.

% , showing superiority in real-world VAD scenarios without training data.

% Differing from conventional methods, our model focuses on action semantics, eliminating the necessity for VAD source data by training solely on an external action skeleton dataset.

\textbf{Comparison with LVLM-based methods.} With the advancements in Large Vision-Language Models (LVLMs)~\cite{liu2023improvedllava,qwen2.5vl,tang2025tspo,anomaly-onevision}, LAVAD~\cite{zanella2024harnessing} proposes a zero-shot video anomaly detection (ZS-VAD) framework. However, it relies on multi-stage reasoning and the coordination of multiple large models with over \textbf{13 billion (B)} parameters, posing challenges for widespread deployment. In contrast, we develop a lightweight zero-shot anomaly detector with a mere \textbf{5.0 million (M)} parameters, just one in two thousand of LAVAD's parameters.
% or the skeleton-based methods, they exhibit better generalization between UBnormal and ShanghaiTech datasets as the human behavioral patterns in these datasets are more similar. However, for datasets like NWPU and UCF-Crime with more complex normal and abnormal types of behavior, previous methods exhibit poor performance when lacking target domain training data to get an understanding of normality and abnormality.

% \begin{figure*}[t]
% \begin{center}
% \includegraphics[scale=0.55]{vis_1.pdf}
% \end{center}
% \caption{(a) Comparison between our typicality modeling module and the prompt-based schemes. (b)-(c) Comparison of different selection ratios in the typicality knowledge selection step.} \label{fig3}
% \end{figure*}

\begin{table}[t]
\setlength\tabcolsep{3.0pt}
\small
\centering
\begin{tabular}{l|cccc}
  \toprule
\,\,\,\,\,\, Experiments&ShanghaiTech&UBnormal&NWPU&UCF-Crime\\
   % \cline{5-13}{0.1pt}
   % \Xcline{5-13}{0.01pt}
 % \cmidrule(r){5-7} \cmidrule(r){8-10} \cmidrule(r){11-13} 
  % stg-nf+typicality data& \bf{85.6}&69.4&60.4&59.5 \\ 
  %  stg-nf+typicality w/o unique 69.9,57.4,52.2
  % (d) prompt + OoD &-&64.1&61.9&56.7\\
 \midrule
  (a) ours w/o aligning&83.2&69.4&60.4&59.5 \\
(b) ours w/o selection&-&64.1&61.9&56.7 \\
(c) ours w/o NF&83.4&72.2&\bf{62.6}&60.5 \\
 % (d) prompt score&81.3&64.4&61.5&61.0 \\
 % \midrule
(d) ours&\bf{84.1}&\bf{74.5}&62.1&\bf{62.7} \\
   \bottomrule
\end{tabular}
\caption{Ablation experiments of the typicality modeling module.}\label{typical}
\end{table}

\begin{table}[t]
\setlength\tabcolsep{5.5pt}
\small
\centering
\begin{tabular}{c|ccc|cccc}
  \toprule
\multirow{2}{*}{Exp.}&\multirow{2}{*}{\bf{Typicality}}&\multicolumn{2}{c|}{\bf{Uniqueness}}&\multirow{2}{*}{SHT}&\multirow{2}{*}{UB}&\multirow{2}{*}{NWPU}&\multirow{2}{*}{UCF-C}\\
&&Cross&Self&&&&\\
   % \cline{5-13}{0.1pt}
   % \Xcline{5-13}{0.01pt}
 % \cmidrule(r){5-7} \cmidrule(r){8-10} \cmidrule(r){11-13} 
    \midrule
(a)&\checkmark&&& 81.9&73.2&62.1&59.6 \\
% \midrule
(b)&&\checkmark&& 81.9&62.9&60.7&59.9 \\
(c)&&&\checkmark& 67.8&60.1&61.0&62.6 \\
(d)&&\checkmark&\checkmark& 82.0&64.5&61.7&61.0 \\
% \midrule
(e)&\checkmark&\checkmark&\checkmark& \bf{84.1}&\bf{74.5}&\bf{62.1}&\bf{62.7} \\
   \bottomrule
\end{tabular}
\caption{{Ablation study of the uniqueness analysis module and holistic anomaly scoring.}}\label{uniquess}
\end{table}

\subsection{Ablation Study}
\textbf{Ablation of typicality module.}
We conduct ablation experiments on the typicality modeling module with the following settings: (a) removing the aligning stage, training an STG-NF~\cite{stgnf} network with our typicality data; (b) removing the collection phase, training with VAD source data (ShanghaiTech); (c) removing the normalizing flow and calculating typicality scores using k-nearest neighbors distance techniques. As shown in Table~\ref{typical}, the model shows poor performance without the aligning stage, as it fails to learn generalizable and discriminative semantic representations. Moreover, performance deteriorates without the selection of typicality action knowledge, as the model can only learn a limited normality boundary from the VAD source data. Furthermore, without the normalizing flow (NF), the model also loses flexibility in modeling the distribution of typical behaviors.

\textbf{Ablation of uniqueness module.} As demonstrated in Table~\ref{uniquess}, when only using the cross-person distance, the model can identify contextual anomalies with acceptable performance. When combined with the self-inspection score, the model can spot changes in motion states, aiding in detecting a wider range of anomaly categories. 

{\textbf{Ablation of holistic anomaly scoring.} As shown in Table~\ref{uniquess}, we conduct ablation experiments on the settings of "sole typicality" (Exp. (a)), "sole uniqueness" (Exp. (b)), and their joint utilization (Exp. (d)).  Our approach achieves optimal performance by integrating the two modules. Notably, the performance of "sole typicality" is higher than that of "sole uniqueness" on UBnormal. The reason is that UBnormal is a synthetic dataset, in which some videos only contain a single person with relatively short movement durations and thus do not align well with real-world surveillance video scenarios.}

\begin{figure*}[t] % h:当前位置, t:顶部, b:底部, p:独立一页
    \centering
    
    % --- 左侧图 ---
    \begin{minipage}[t]{0.46\textwidth} % 0.48代表占据整行48%的宽度
        \centering
        % 注意：这里的 \textwidth 指的是当前 minipage 的宽度
        \includegraphics[width=\textwidth]{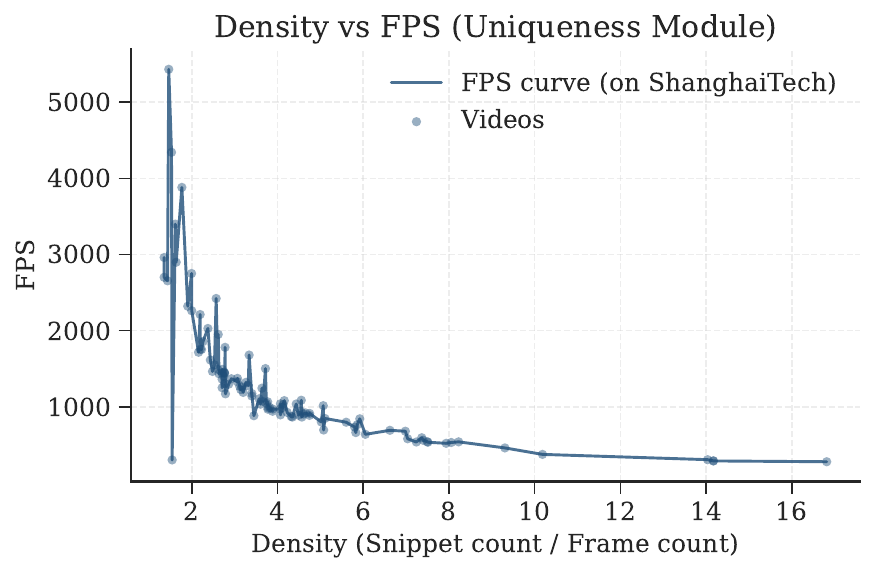} 
        \caption{{Uniqueness module runtime.}} 
        \label{fig:line_chart} % 给左图一个独立的标签
    \end{minipage}
    \hfill % \hfill 会自动撑开左右两张图之间的空白
    % --- 右侧图 ---
    \begin{minipage}[t]{0.46\textwidth}
        \centering
        \includegraphics[width=\textwidth]{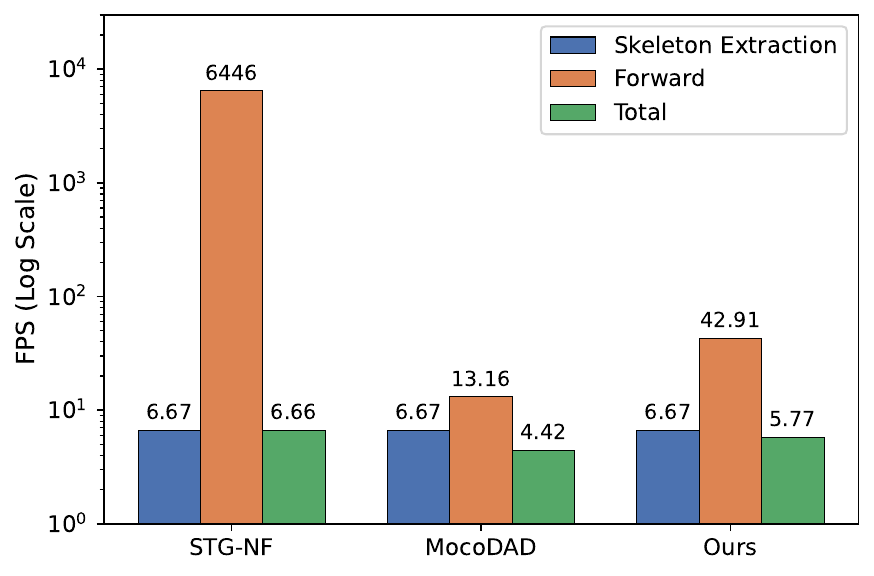} 
        \caption{{Detailed runtime comparison.}} 
        \label{fig:bar_chart} % 给右图一个独立的标签
    \end{minipage}
    
\end{figure*}

\textbf{Runtime analysis.}
 {We conduct a detailed analysis of the efficiency of our method. The computational cost of the test-time uniqueness module mainly lies in KNN calculation. The complexity of KNN distance calculation is ${O(qN)}$, where $q$ denotes the number of trajectory snippets for a query person and $N$ represents the number of pedestrian trajectory snippets in a single video. As shown in Fig.~\ref{fig:line_chart}, the FPS of the uniqueness module decreases in more densely populated pedestrian scenes, owing to the increased complexity of KNN calculation. \textbf{Nevertheless}, the processing time of this module is negligible, because the computational bottleneck of skeleton-based methods lies in the extraction of skeleton trajectories — as illustrated in Fig.~\ref{fig:bar_chart}, this step only achieves 6.67 FPS.
Compared with STG-NF, our model has a longer forward time, since the forward process of our model incorporates skeleton feature extraction (45.12 FPS) as well as the calculation of typicality and uniqueness (876 FPS). Even so, there is an insignificant difference in the total FPS. In addition, our method achieves a 3× faster model forward speed than MocoDAD while delivering significantly superior performance (see Table~\ref{tab1}). }

% \begin{table}[t]
% \setlength\tabcolsep{6.5pt}
% \small
% \centering
% \caption{Ablation results of the nearest neighborhood
% (NN) number $k$ and the masking threshold $\alpha$ in the uniqueness analysis
% module.}\label{hyper-s}
% \begin{tabular}{ccc|ccc}
%   \toprule
% ($k,\alpha$)&ShanghaiTech&UB&($k,\alpha$)&ShanghaiTech&UB\\
%  \midrule
%  (1,\;1)&83.9&74.2& (16,\;1)&\bf{84.2}&74.2 \\
%  (4,\;1)&84.1&74.3 & (16,\;2)&84.2&74.4\\
%  (16,\;1)&\bf{84.1}&74.5 &(16,\;3)&84.0&74.5\\
%  (64,\;1)&83.7&\bf{74.7}&(16,\;4)&84.1&\bf{74.5} \\
%    \bottomrule
% \end{tabular}
% \end{table}

% \begin{table}[t]
% \setlength\tabcolsep{6.5pt}
% \caption{Hyper-parameter sensitivity of the selection ratio.}\label{hyper}
% \small
% \centering
% \begin{tabular}{ccc|ccc}
%   \toprule
% ($\beta^n,\beta^a$)&ShanghaiTech&UB&($\beta^n,\beta^a$)&ShanghaiTech&UB\\
%  \midrule
%  (0.1,\;0.1)&82.8&68.3& (0.9,\;0.1)&\bf{84.1}&\bf{74.5} \\
%  (0.3,\;0.1)&83.0&70.9 & (0.9,\;0.3)&84.1&73.3\\
%  (0.5,\;0.1)&83.4&74.1 &(0.9,\;0.5)&84.1&72.3\\
%  (0.7,\;0.1)&83.9&74.3&(0.9,\;0.7)&83.5&72.0 \\
%  (0.9,\;0.1)&\bf{84.1}&\bf{74.5} &(0.9,\;0.9)&82.5&71.8\\
%    \bottomrule
% \end{tabular}
% \end{table}

\textbf{Comparison with prompt-based methods.} \label{ablate}
Since prompt-based techniques have been popular in other zero-shot tasks~\cite{sato2023prompt,winclip,li2024promptad}, we conduct experiments to compare our typicality module with theirs. To this end, we design typical normal prompts, typical abnormal prompts, and ensemble prompts, and then use skeleton-prompt similarity as the anomaly score. Specifically, we adopt a normal prompt list: [``usual'', ``normal'', ``daily'', ``stable'', ``safe''], and an abnormal prompt list: [``danger'', ``violence'', ``suddenness'', ``unusual'', ``instability'']. The prompts are encoded into text features and compute similarity with the skeleton features, together with our uniqueness analysis. As shown in Table~\ref{prompt}, the results are suboptimal. Unlike various forms of text seen in CLIP image-text alignment, the current skeleton-text alignment scheme has only encountered text of action class names, thus the alignment capability for prompt text is weak. Our method, on the other hand, focus on semantic typicality and context uniqueness, avoiding directly using the skeleton-prompt similarity.

 % we design typical normal Prompts (NP), such as "usual", "stable", and typical abnormal prompts (AP), such as "violence", "fighting", and ensemble two types of prompts (NAP),

    \begin{table}[t]
\setlength\tabcolsep{5.0pt}
\small
\centering
\begin{tabular}{l|cccc}
  \toprule
\,\,\,\,\,\, Experiments&ShanghaiTech&UBnormal&NWPU&UCF-Crime\\
   % \cline{5-13}{0.1pt}
   % \Xcline{5-13}{0.01pt}
 % \cmidrule(r){5-7} \cmidrule(r){8-10} \cmidrule(r){11-13} 
  % stg-nf+typicality data& \bf{85.6}&69.4&60.4&59.5 \\ 
  %  stg-nf+typicality w/o unique 69.9,57.4,52.2
  % (d) prompt + OoD &-&64.1&61.9&56.7\\
 \midrule
  (a) normal prompts&81.3&64.4&61.5&61.0 \\
(b) abnormal prompts&80.7&63.6&61.0&60.2 \\
(c) ensemble prompts&80.7&63.9&61.1&61.3 \\
(d) ours&\bf{84.1}&\bf{74.5}&\bf{62.1}&\bf{62.7} \\
   \bottomrule
\end{tabular}
\caption{Comparison with prompt-based zero-shot methods.}\label{prompt}
\end{table}

    % \begin{subfigure}[b]{0.885\textwidth}
    %     \centering
    % \includegraphics[width=\textwidth]{vis_main_2.pdf} 
    %     \caption{Unique anomalies Localization. (Left) STG-NF, (Right) Ours.}
    %     \label{fig:sub2}
    % \end{subfigure}

\begin{table}[t]
\centering
\small
\setlength\tabcolsep{6.0pt}
\begin{tabular}{lcccc} % 文本左对齐(l)，数字居中(c)
  \toprule
  Method & ShanghaiTech & UBnormal & NWPU & UCF-Crime \\
  \midrule
 GPT-3.5-Turbo& 84.1  & 74.5 & 62.1 & 62.7 \\
 Gemini-3-Pro& 84.6 & 74.2 & 61.4 & 62.5 \\
 GPT-5.4-mini&84.7&74.5&61.5&62.0\\
 Identical &84.8&73.3&61.6& 60.2 \\
  \bottomrule
\end{tabular}
\caption{{Sensitivity analysis with different LLMs in the typicality selection phase.}}\label{llm}
\end{table}

{\textbf{Ablation of LLMs. } To investigate the robustness of our proposed method to different LLMs, we conduct validations and identical test on three LLMs, all using their official clients and default parameter settings, as shown in Table~\ref{llm}. Although the output results of different LLMs exhibit slight discrepancies, our proposed model still demonstrates favorable robustness. The underlying reasons are mainly threefold: \textbf{(1) Consistency of empirical cognition}: The normal actions labeled by all models are daily, low-activity behaviors (e.g., reading a book, walking the dog), while the abnormal actions are all intense activities with large movement amplitudes (e.g., skydiving, pushing person); \textbf{(2) Robustness support from Normalizing Flow}: We adopt Normalizing Flow as a robust binary classifier, which projects input features onto the standard normal distribution. This makes features with higher occurrence frequencies locate at the center of the distribution and possess the highest probabilities, thus filtering out the impact of a small number of noisy labels. \textbf{(3) Uniqueness-based perspective}: We also introduce a test-time uniqueness module, which provides an alternative evidence for anomaly detection from another perspective and further enhances the model robustness.
In addition, we carry out a identical test experiment: we screen out the typical normal/abnormal actions consistently identified by all models, and ultimately obtain 5 typical normal actions and 4 typical abnormal actions. However, this experiment does not yield optimal results, due to the reduced number of typical samples after screening, which prevents the model from sufficiently learning diverse sample features during the training phase.}

\begin{table}[t]
\setlength\tabcolsep{4pt}
\small
\centering
\begin{tabular}{c|c|c|c}
  \toprule
Method&Backbone&Params.&Performance\\
 \midrule
Jigsaw~\cite{jigsaw}&3D-Conv&1.5M& 58.6\,/\, 61.1\,/\, 53.3\\
HF2-VAD~\cite{hf2}&MemAE+C-VAE&36.4M&59.5\,/\,58.3\,/\,52.9\\
STG-NF~\cite{hf2}&STG-NF&0.7K&68.8 \,/\,57.6 \,/\,51.6\\
\midrule
\multirow{3}*{Ours}&STG-NF&0.7K& 69.0\,/\,60.1\,/\,58.5\\
&ST-GCN\,+\,NF&4.1M&73.1\,/\,61.3\,/\,61.5 \\
&CTR-GCN\,+\,NF&5.0M&\bf{74.5} \,/\,\bf{62.1}\,/\, \bf{62.7}\\
   \bottomrule
\end{tabular}
\caption{Ablation studies of the backbone networks. The performance is evaluated on UBnormal\,/\,NWPU\,/\,UCF-Crime datasets.}\label{backbone}
\end{table}

\textbf{Ablation of backbone. }In this part, we ablate our backbone. For a fair comparison with our baseline STG-NF~\cite{stgnf}, we attempt to use STG-NF as the backbone. However, STG-NF takes $XY$-coordinates as input, with only 2 dimensions, making it extremely lightweight yet difficult to learn high-level features. We then use STG-NF to learn the typicality skeleton coordinate inputs to obtain the typicality score. (Note that the results in Table~\ref{typical} (a) of the main paper also consider the uniqueness score, resulting in the higher performance). As shown in Table~\ref{backbone}, using STG-NF as our backbone still demonstrates better performance compared to vanilla STG-NF, highlighting the effectiveness of our typicality training. In addition, when we switch our backbone from CTR-GCN~\cite{ctr} to {ST-GCN}~\cite{stgcn}, the model becomes more lightweight and the performance remains good.

% Thus, we can conclude that our method is a computationally efficient and performance-superior approach for skeleton-based ZS-VAD task.

\begin{figure*}[t]
\begin{center}
\includegraphics[scale=0.35]{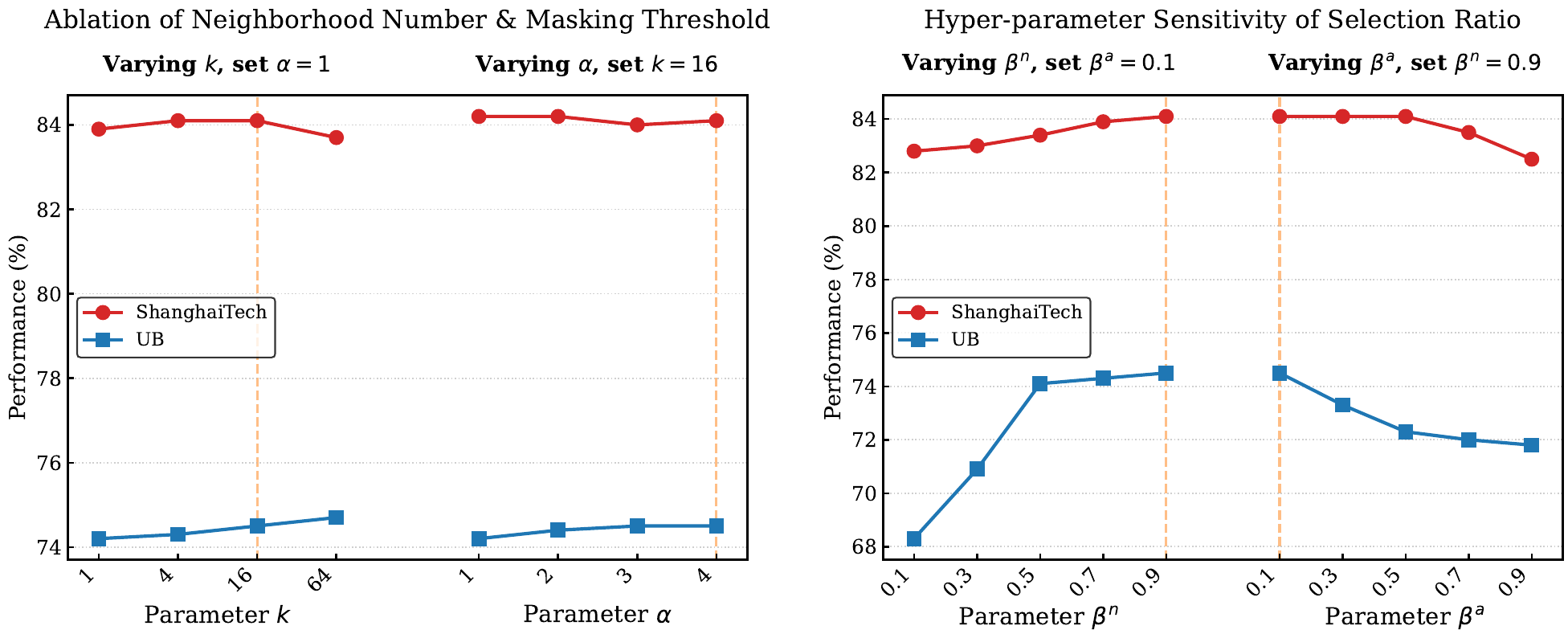}
\end{center}
\caption{{Sensitivity analysis of hyper-parameters: neighborhood number ($k$), masking threshold ($\alpha$), and selection ratios ($\beta^n$, $\beta^a$).}} \label{hyperp}
\end{figure*}

\textbf{Hyper-parameter analysis}. We ablate the nearest neighborhood (NN) number $k$ and the masking threshold $\alpha$ in the uniqueness analysis module. As shown in Fig.~\ref{hyperp}, our method is robust for these two hyper-parameters. Choosing an appropriate $k$ can \textbf{filter out some unrelated activities} and focus solely on behaviors related to the current skeleton snippets. In addition, taking the average of the $k$ neighbors helps suppress noise, which also makes our model insensitive to $\alpha$. The optimal values of k and $\alpha$ are 16 and 4, respectively.

We also ablate the hyperparameters in the typicality knowledge selection step. The optimal values of $\beta^n$ and $\beta^a$ are 0.9 (90\%) and 0.1 (10\%), respectively. A smaller $\beta^a$ can enhance performance by \textbf{filtering out noisy data and normal snippets within the anomalous sequences}.

\begin{figure*}[t]
    \centering
    % 第一行
    % \begin{subfigure}[b]{0.9\textwidth}
    %     \centering
        \includegraphics[width=\textwidth]{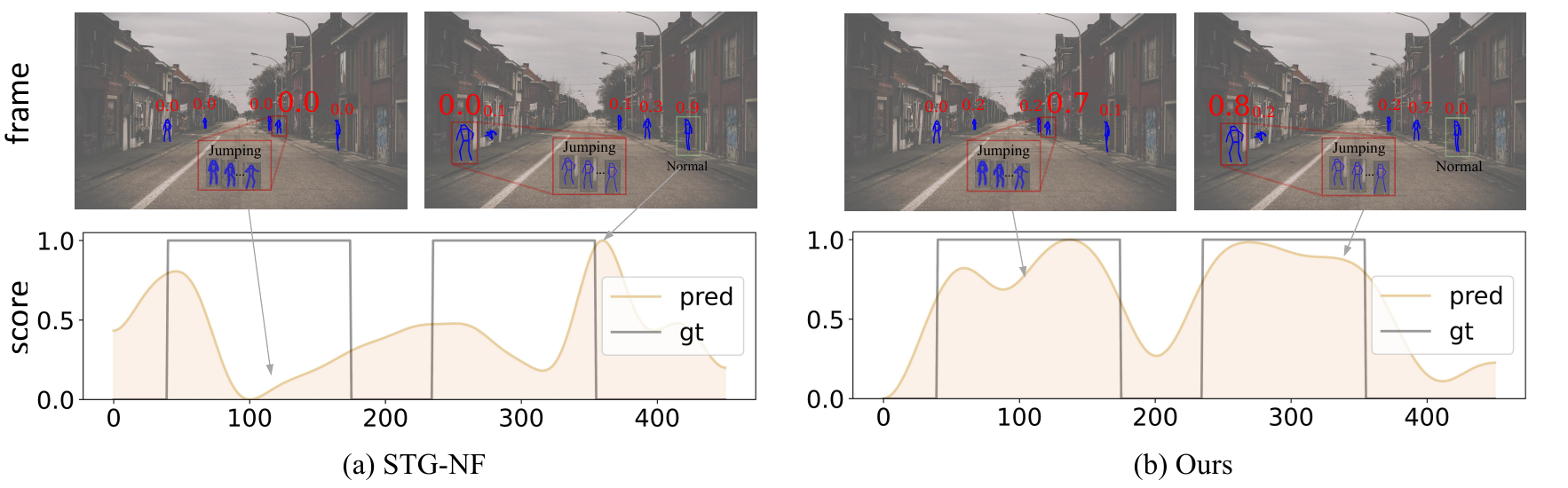} 
        % \caption{Typical anomalies Localization. (Left) STG-NF, (Right) Ours.}
        % \label{fig:sub1}
    % \end{subfigure}
    % \vspace{0.5cm} % 图之间的垂直间距
    % 第二行
    \caption{{Example results of our method that succeed in capturing \textbf{typical anomalies}. STG-NF~\cite{stgnf} fails to detect the ``jumping in the street'' event, while ours performs well through action typicality learning.  Each individual ({blue skeleton}) has a predicted anomaly score (\textcolor{red}{red font}), where the frame-level score (\textcolor{orange}{orange line}) is defined as the maximum among all individuals in that frame. For clearer visualization, we present the action sequence of ``jumping in the street'' in the figure.}}  
    \label{vis_main}
\end{figure*}
% Hyper-parameter sensitivity. There are two main hyper-parameters: the ratio of anomaly skeleton
% data β, and the number of nearest neighborhood (NN) numbers k. As shown in Fig. 3 (b), a smaller
% beta can lead to better performance as it helps filter out normal segments within the anomalous data.
% As shown in Fig. 3 (c), our model demonstrates robustness concerning the NN number.
% with the following settings: (d) only using the cross-person score, (e) adding the self-inspection score, and (f) incorporating both cross-person and self-person score, \emph{i.e.}, the uniqueness score. 

% 在仅使用唯一性评分在UBnormal数据集上表现不佳的原因是，UBnormal是一个合成数据集，其中一些视频仅包含一个人，运动持续时间相对较短，这与真实监控视频场景不太吻合。

% \begin{figure}[t]
% \begin{center}
% \includegraphics[scale=0.5]{ICLR 2025 Template/imgs/vis_main2.pdf}
% \end{center}
% \caption{Example results of our method that succeeds in capturing unique anomalies under ZS-VAD setting. For the ``photographing at restricted areas" anomaly occurring between approximately frames 2600 to 3000, STG-NF~\cite{stgcf} fails to detect it, whereas our method can detect it through context uniqueness analysis. } \label{fig5}
% \end{figure}

\subsection{Visualization Results}
Fig. \ref{vis_main} presents the qualitative localization results of typical anomalies.  {The skeleton movement of jumping are different from those of normal walking, such as the larger swing of the arms and the greater upward lift of the thighs.} STG-NF fails to detect the ``jumping'' anomaly due to its low-level skeleton representation, which is unable to distinguish between indiscernible anomaly patterns that are similar to normal patterns. Disturbed by the skeletal noise of frame 350, it erroneously identifies the anomaly's position. In contrast, our model maps the skeleton snippets to a high-level space with generalizable and discriminative semantic information, allowing it to identify the anomaly based on our trained decision boundary. 
In surveillance scenarios lacking training samples, our model can still be effectively utilized to detect certain typical abnormal behaviors.

\begin{figure*}[t]
    \centering
    % 第一行
    % \begin{subfigure}[b]{0.9\textwidth}
    %     \centering
\includegraphics[width=0.9\textwidth]{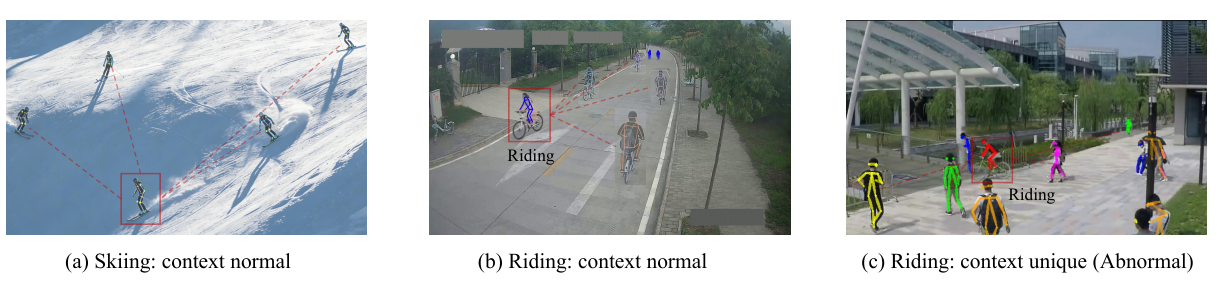} 
        % \caption{Typical anomalies Localization. (Left) STG-NF, (Right) Ours.}
        % \label{fig:sub1}
    % \end{subfigure}
    % \vspace{0.5cm} % 图之间的垂直间距
    % 第二行
    \caption{{Visual illustration of how the contextual uniqueness analysis module works in different scenes. For actions that cannot be judged by the typicality anomaly score, the uniqueness score can serve as a calibration to improve the reliability of the anomaly score.}} 
    \label{ski}
\end{figure*}

%
% \textbf{Unique Anomaly Localization.}
{The uniqueness module serves as a complement to the typicality prior. 
As shown in Fig.~\ref{ski} (a) and (b), for specific scenarios such as ski resorts and bicycle lanes, similar actions are exhibited by different individuals across various time periods. The high behavioral similarity among different individuals results in a low uniqueness score for such actions. In addition, the same action may yield different classification results in distinct contexts. As illustrated in Fig.~\ref{ski} (c), riding in this scene differs significantly from the walking behaviors of surrounding individuals, leading to a high uniqueness score for the action.}

% \clearpage
% \setcounter{page}{1}
% \maketitlesupplementary

\begin{figure*}[t]
    \centering
    % 第一行
    % \begin{subfigure}[b]{0.9\textwidth}
    %     \centering
        \includegraphics[width=\textwidth]{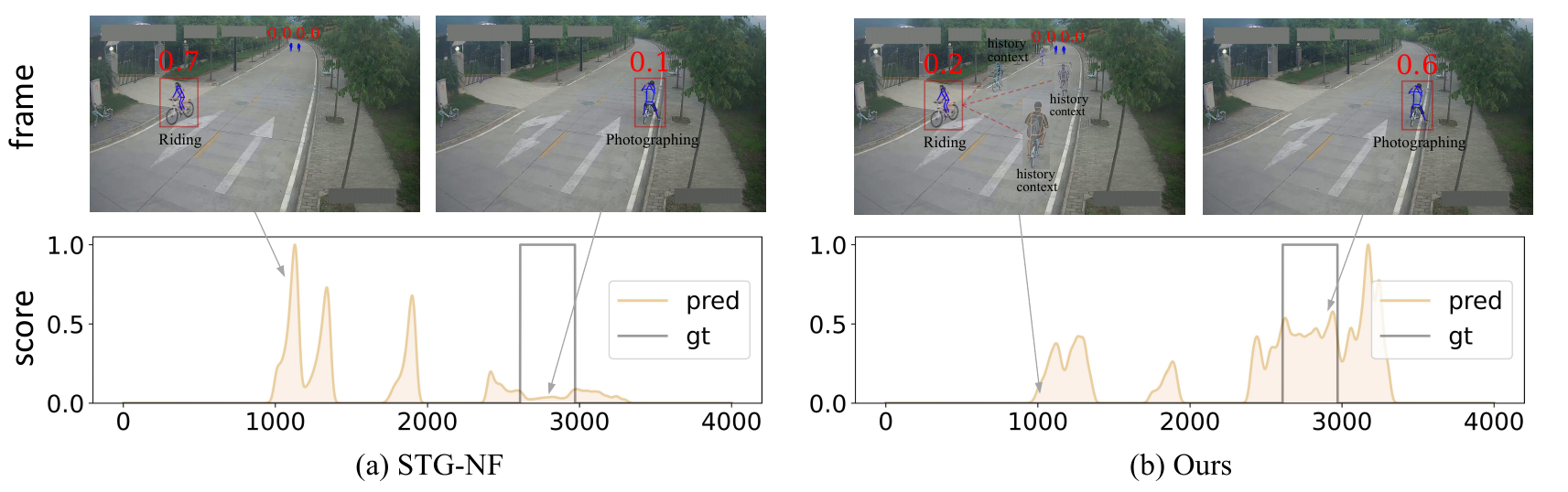} 
        % \caption{Typical anomalies Localization. (Left) STG-NF, (Right) Ours.}
        % \label{fig:sub1}
    % \end{subfigure}
    % \vspace{0.5cm} % 图之间的垂直间距
    % 第二行
    \caption{{Example results of our method that succeed in capturing unique anomalies. STG-NF  misclassifies unseen normal events during periods where “riding” occurs as anomalies, whereas our method correctly identifies them as normal by recognizing their contextual similarity. Moreover, STG-NF fails to detect the “photographing in restricted areas” anomaly, while our approach successfully identifies it by recognizing a sudden change in the person's trajectory.}} 
    \label{vis_main2}
\end{figure*}

% \section{Codes}\label{code}
% The inference code and pre-trained model weights are included in the supplementary materials for evaluation.
 Fig. \ref{vis_main2} shows the qualitative localization results of unique anomalies.  Existing skeleton-based methods rely on the source normal data for training. When the source domain does not include some novel behavior that appears in the target domain, these behavior will be classified as anomalies. Consequently, STG-NF erroneously localizes the anomaly during time periods when ``riding'' is present.  In contrast, our model can analyze the spatio-temporal differences and establish scene-adaptive decision boundaries. Since ``riding'' occurs multiple times in the video, its uniqueness score is low. On the other hand, ``photographing at restricted areas'' exhibits significant differences from the surrounding people's behavior and appears as a sudden change in the person's movement trajectory, resulting in a corresponding increase in its anomaly score.

\section{Conclusion}~\label{conclusion}
In this paper, we identify the advantages of the skeleton-based approach in zero-shot video anomaly detection, and introduce a novel framework that can generalize to various target scenes with semantic typicality and context uniqueness learning. First, we propose a language-guided typicality modeling module that effectively learns the typical distribution of normal and abnormal behavior. Secondly, we propose a test-time context uniqueness analysis module to derive scene-adaptive boundaries. Comprehensive experiments across four large-scale VAD datasets demonstrate
the effectiveness and generalizability of our model. {\textbf{Limitations and future work}: This work focuses on unlocking the full potential of skeleton-based methods for ZS-VAD, yet the scene information is less exploited. Future work will integrate the joint understanding of both  actions and scene within the zero-shot setting.}

\section*{Acknowledgment}
This work was supported in part by the National Science and Technology Major Project under Grant 2023ZD0121300, in part by the National Natural Science Foundation of China under Grant 62088102, Grant U24A20325, and Grant 12326608, and in part by the Fundamental Research Funds for the Central Universities under Grant XTR042021005.

\bibliographystyle{elsarticle-num}

\bibliography{main.bib}

\end{document}